\documentclass[11pt,a4paper]{article}
\usepackage[hyperref]{naaclhlt2019}
\usepackage{times}
\usepackage{url}
\usepackage{latexsym}
\usepackage{graphicx}
\usepackage{multirow}
\usepackage{color}
\usepackage{pgfplots}
\usepackage{amsmath}
\usepackage{amssymb}
\usepackage[section]{placeins}
\usepackage[colorinlistoftodos]{todonotes}
\usepackage{booktabs}

\global \hyphenpenalty 2000
\global \exhyphenpenalty 1000

\newcommand{\svdpmi}{SVD$_{\text{PPMI}}$}
\newcommand\blfootnote[1]{%
  \begingroup
  \renewcommand\thefootnote{}\footnote{#1}%
  \addtocounter{footnote}{-1}%
  \endgroup
}

\newcommand{\myurl}[1]{\href{http://#1}{\tt \nolinkurl{#1}}}

\renewcommand{\sim}{\mathrm{sim}}
\newcommand{\pos}{\mathrm{POS}}
\renewcommand{\neg}{\mathrm{NEG}}

\aclfinalcopy 

\begin{document}
\title{Modeling Word Emotion in Historical Language: \\ 
Quantity Beats Supposed Stability in Seed Word Selection}

\author{Johannes Hellrich*    \hspace{25pt}  Sven Buechel*    \hspace{25pt} Udo Hahn \vspace{8pt}\\
\{\texttt{firstname.lastname}\}\texttt{@uni-jena.de} \vspace{4pt} \\
Jena University Language \& Information Engineering (JULIE) Lab\\
Friedrich-Schiller-Universit\"at Jena, Jena, Germany\\
\myurl{julielab.de}
}
\maketitle
\blfootnote{* These authors contributed equally to this work. Johannes Hellrich was responsible for selecting historical text corpora and training embedding models. Sven Buechel selected existing emotion lexicons and was responsible for modeling word emotions. The adaptation of polarity-based algorithms (Section \ref{sec:method}), the creation of the German and English historical gold standard lexicons (Section \ref{sec:gold}), as well as the overall study design were done jointly.}
 
\begin{abstract}
To understand historical texts, we must be aware that language---including the emotional connotation attached to words---changes over time. In this paper, we aim at estimating the emotion which is associated with a given word in former language stages of English and German. Emotion is represented following the popular Valence-Arousal-Dominance (VAD) annotation scheme. While being more expressive than polarity alone, existing word emotion induction methods are typically not suited for addressing it. To overcome this limitation, we present adaptations of two popular algorithms to VAD. To measure their effectiveness in diachronic settings, we present the first gold standard for historical word emotions, which was created by scholars with proficiency in the respective language stages and covers both English and German. In contrast to claims in previous work, our findings indicate that hand-selecting small sets of seed words with supposedly stable emotional meaning is actually harm- rather than helpful.
\end{abstract}
\section{Introduction}

Language change is ubiquitous and, perhaps, most evident in lexical semantics. In this work, we focus on changes in the affective meaning of words over time.
Although this problem has been occasionally addressed in previous work (see Section \ref{subsec:historicalSentiment}), most contributions in this area are limited to a rather shallow understanding of human emotion, typically in terms of \textit{semantic polarity} (feelings being either positive, negative or neutral). 
Another major shortcoming of this area is the lack of appropriate data and methodologies for evaluation. As a result, the aptness of  algorithmic contributions has so far only been assessed in terms of face validity rather than quantitative performance figures  \citep{Cook10,Buechel16lt4dh,Hamilton16emnlp,Hellrich18coling}.

To tackle those shortcomings, we first introduce adaptations of algorithms for word polarity induction to vectorial emotion annotation formats, thus enabling a more fine-grained analysis. Second, to put the evaluation of these methods on safer ground, we present two datasets of affective word ratings for English and German, respectively.\footnote{ Publicly available together with experimental code at\\
\mbox{\myurl{github.com/ JULIELab/HistEmo}}} These have been annotated by scholars in terms of language-stage-specific emotional connotations.

We ran synchronic as well as diachronic experiments to compare different algorithms for modeling historical word emotions---the latter kind of evaluation employs our newly created gold standard. 
In particular, one prominent claim from previous work has been that \textit{full-sized} emotion lexicons of contemporary language are ill-suited for inducing historical word emotion. Rather, it would be much more beneficial to select a small, \textit{limited} set of seed words of supposedly invariant emotional meaning \cite{Hamilton16emnlp}. In contrast, our experiments indicate that larger sets of seed words perform better than manually selected ones despite the fact that some of their entries may not be accurate for the target language stage. 
Our unique historical gold standard is thus an important step towards firmer methodological underpinnings for the computational analysis of textually encoded historical emotions.

\section{Related Work}
\label{sec:related}

\subsection{Representing Word Emotions} \label{sec:representing}

\begin{figure}[t]
    \includegraphics[width=0.48\textwidth]{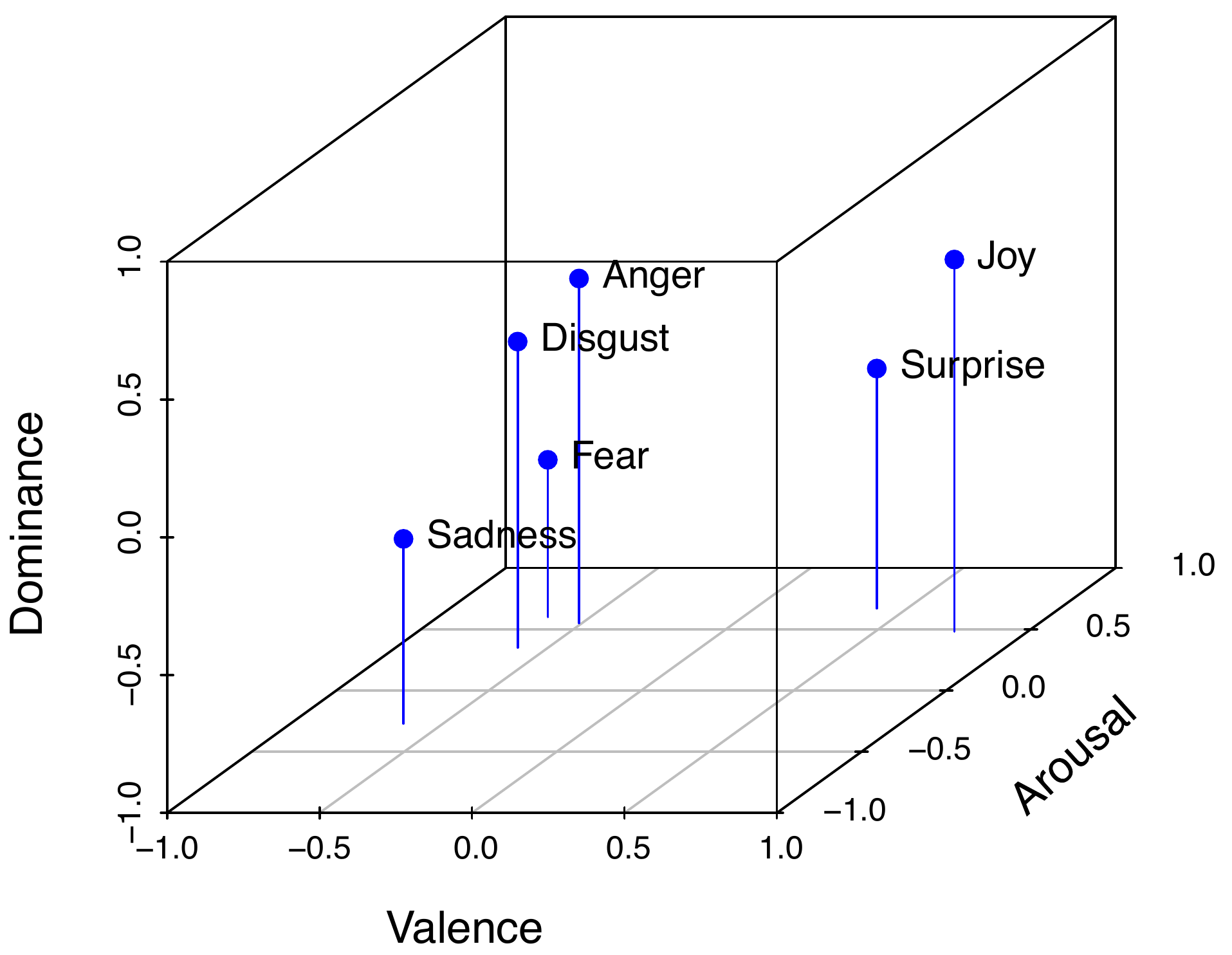}
\caption{\label{fig:VAD} 
Affective space spanned by the Valence-Arousal-Dominance (VAD) model, together with the position of six basic emotion categories.}
\end{figure}

Quantitative models for word emotions can be traced back at least to  \citet{Osgood53} who used questionnaires to gather human ratings for words on a wide variety of dimensional axes including 
``\textit{good} vs. \textit{bad}''. Most previous work focused on varieties of such forms of semantic polarity, a rather simplified representation of the richness of human affective states---an observation increasingly recognized in sentiment analysis \cite{Strapparava16}. In contrast to this bi-polar representation, the Valence-Arousal-Dominance (VAD) model of emotion \cite{Bradley94} is a well-established approach in psychology \cite{Sander09} which increasingly attracts interest by NLP researchers \citep{Koeper16,Yu16,Wang16,Shaikh16,Buechel17eacl,Preotiuc16,Mohammad18acl}. 
The VAD model assumes that  affective states can be characterized relative to Valence (corresponding to the concept of polarity), Arousal (the degree of calmness or excitement) and Dominance (perceived degree of control). Formally, VAD spans a three-dimensional real-valued space  (see Figure \ref{fig:VAD}) making the prediction of such values a multi-variate regression problem \cite{Buechel16ecai}.

Another popular line of emotion representation evolved around the notion of \textit{basic emotions}, small sets of discrete, cross-culturally universal affective states \citep{Scherer00}. Here, contributions most influential for NLP are Ekman's \citeyearpar{Ekman92} six basic emotions as well as Plutchik's \citeyearpar{Plutchik80} wheel of emotion \citep{Strapparava07,Mohammad13,Bostan18coling}. In order to illustrate the relationship between Ekman's basic emotions and the VAD affect space the former are embedded into the latter scheme in Figure \ref{fig:VAD}.

The affective meaning of individual words is encoded in so-called \textit{emotion lexicons}.
Thanks to over two decades of efforts from psychologists and AI researchers alike, today a rich collection of empirically founded emotion lexicons is available covering both VAD and basic emotion representation for many languages (see \citet{Buechel18lrec} for an overview). One of the best know resources of this kind are the \textit{Affective Norms for English Words} (\textsc{Anew}; \citealp{Bradley99}) which comprise 1,034 entries in VAD format. \textsc{Anew}'s popular extension by \citet{Warriner13} comprises roughly 14k entries acquired via crowdsourcing (see Table \ref{tab:examples} for examples).

Recently, researchers started to build computational models of the relationship between VAD and discrete categories (illustrated in Figure \ref{fig:VAD}) resulting in techniques to automatically translate ratings between these major representation schemes \cite{Calvo13,Buechel18coling}. 

\begin{table}[tb]
\small
\center
\begin{tabular}{l|ccc}
\toprule
Entry & Valence & Arousal & Dominance\\
\midrule
\textit{rage}	&2.50	&6.62	&4.17\\
\textit{orgasm}&	8.01	&7.19	&5.84\\
\textit{relaxed}	&7.25	&2.49	&7.09\\
\bottomrule
\end{tabular}
\caption{\label{tab:examples} 
Sample Valence-Arousal-Dominance (VAD) ratings from the emotion lexicon by \newcite{Warriner13}. The scales span the interval of $[1,9]$ for each dimension, ``$5$'' being the neutral value.}	
\end{table}

\subsection{Predicting Word Emotions} 
Word emotion induction---the task of predicting the affective score of unrated words
---is an active research area within sentiment analysis \citep{Rosenthal15}. Most approaches either rely on hand-coded lexical resources, such as \textsc{WordNet} \cite{Fellbaum98}, to propagate sentiment information to unkown words \citep{Shaikh16}, or employ similarity metrics based on distributional semantics (see below). We deem the former inadequate for diachronic purposes, since almost all lexical resources typically cover contemporary language only. In the following, we focus on algorithms which have been tested in diachronic settings in previous work. An overview of recent work focusing on applications to contemporary language is 
given by \citet{Buechel18naacl}.

More than a decade ago, \newcite{Turney03} introduced a frequently used and often adopted (e.g., \citet{Koeper16,Palogiannidi16}) algorithm. It computes a sentiment score based on the similarity of an unrated word to two sets of positive and negative seed words.
\newcite{Bestgen08} presented an algorithm which has been prominently put into practice in expanding a VAD lexicon to up to 17,350 entries \cite{Bestgen12}. Their method employs a k-Nearest-Neighbor methodology where an unrated word inherits the averaged ratings of the surrounding words.
\newcite{Rothe16} presented a more recent approach to polarity induction. Based on word embeddings and a set of positive and negative paradigm words, they train an orthogonal transformation of the embedding space so that the encoded polarity information is concentrated in a single vector component whose value then serves as an explicit polarity rating. 
The algorithm proposed by \newcite{Hamilton16emnlp} employs a random walk within a lexical graph constructed using word similarities. They outperform \newcite{Rothe16} when embeddings are trained on small datasets. 

Note that these algorithms differ in the kind of input representation they require. Whereas \newcite{Turney03}, \newcite{Rothe16},  and \newcite{Hamilton16emnlp} expect binary class ratings (positive or negative), Bestgen's algorithm \cite{Bestgen08} takes vectorial seed ratings, illustrated in Table \ref{tab:examples}, as input.

\subsection{Historical Sentiment Information}
\label{subsec:historicalSentiment}
There are several studies using contemporary word emotion information, i.e., emotion lexicons encoding today's emotional meaning,  to analyze historical documents. For instance, \citet{Acerbi13} and \citet{Bentley14} observed long-term trends in words expressing emotions in the Google Books corpus and linked these to historical (economic) events.
Another example are \citet{Kim17} who investigate emotions in literary texts in search for genre-specific patterns. However, this contemporary emotion information could lead to artifacts, since the emotions connected with a word are not necessarily static over time. This phenomenon is known as elevation \& degeneration in historical linguistics, e.g., Old English \textit{cniht} `boy, servant' was elevated becoming the modern \textit{knight} \citep{Bloomfield84}.

Alternatively, algorithms for bootstrapping word emotion information can be used to predict historical emotion values by using word similarity based on historical texts. This was first done for polarity regression with the \citet{Turney03} algorithm and a collection of three British English corpora by \citet{Cook10}. \citet{Jatowt14} tracked the emotional development of words by averaging the polarity of the words they co-occurred with (assuming the latters' polarity to be stable). \citet{Hamilton16emnlp} used their novel random walk-based algorithm for polarity regression on COHA. They consider their method especially suited for historical applications.\footnote{
However, the algorithm is sensitive to changes in its training material and thus likely prone to compute artifacts, see their README at \myurl{github.com/williamleif/socialsent}}
This algorithm was also used by \citet{Genereux17} to test the temporal validity of inferred word abstractness, a psychological measure akin to the individual VAD dimensions. They used both modern and historical (1960s) psychological datasets rating the same words as gold standards and found a strong correlation with predicted historical abstractness. \citet{Buechel16lt4dh} used \citet{Bestgen08}'s algorithm to investigate emotional profiles of different genres of historical writing. 
Finally, we used the \citet{Turney03} algorithm to induce historical sentiment information which is provided as part of \myurl{JeSemE.org}, a website for exploring semantic change in multiple diachronic corpora \citep{Hellrich18coling}.

\section{Methods}
\label{sec:method}
\label{sec:algorithms} 

\subsection{Word Similarity}

We measure word similarity by the cosine between word embeddings, the most recent method in studies of distributional semantics.
Their most popular form are Skip-Gram Negative Sampling (SGNS; \citealp{Mikolov13iclr}) embeddings which are trained with a very shallow artificial neural network. SGNS processes one word-context pair, i.e., two nearby words, at a time and learns good embeddings by trying to predict the most likely contexts for a given word. 

An alternative solution for generating low dimensional vectors is gathering all word-context pairs for a corpus in a large matrix and reducing its dimensionality with singular value decomposition (SVD), a technique very popular in the early 1990's \cite{Deerwester90,Schutze93acl}. \newcite{Levy15} propose \svdpmi , a state-of-the-art algorithm based on combining SVD with the positive pointwise mutual information (PPMI; \citealp{Niwa94}) word association metric.

Both SGNS and \svdpmi\ have been shown to be adequate for exploring historical semantics \cite{Hamilton16acl,Hamilton16emnlp}. A general downside of existing embedding algorithms other than \svdpmi\ is their inherent stochastic behavior during training which makes the resulting embedding models unreliable
\cite{Hellrich16coling,Antoniak18,Wendlandt18}. 
Very recently, contextualized word embeddings, such as ELMo \citep{Peters18naacl} and BERT \citep{Devlin18}, have started to establish themselves as a new family of algorithms for word representation. Those methods achieve enhanced performance on many downstream tasks by taking context into account, both during training and testing, to generate an individual vector representation for each individual \textit{token}. This makes them unsuitable for our contribution, since we address emotion on the  \textit{type} level by creating emotion lexicons.  

\subsection{Word Emotion} 
\label{subsec:algos}
Our work employs three algorithms for inducing emotion lexicons, two of which had to be adapted to deal with the more informative vectorial VAD representation instead of a simple binary two-class representation (positive vs. negative polarity):

\begin{description}

\item[\textsc{kNN}]--- The k-Nearest-Neighbor-based al\-go\-rithm by \newcite{Bestgen08} which already supports vectorial input.

\item[\textsc{ParaSimNum}]--- An adaptation of the classical \textsc{ParaSim} algorithm by \newcite{Turney03} which is based on the similarity of two opposing sets of paradigm words.

\item[\textsc{RandomWalkNum}]--- An adaptation of the \textsc{RandomWalk} algorithm proposed by \newcite{Hamilton16emnlp} which propagates affective information of seed words via a random walk through a lexical graph.
\end{description}

\textsc{kNN} sets the emotion values of each word $w$ to the average of the emotion values of the $k$ most similar seed words. For any given seed word $s$, let $e(s)$ denote its three-dimensional emotion vector corresponding to its VAD value in our seed lexicon. Furthermore, let $\mathrm{nearest}(w,k)$ denote the set of the $k$ seed word most similar to a given word $w$. Then the predicted emotion of word $w$ according to \textsc{kNN} is defined as follows:
\begin{equation}
e_\text{\sc kNN}(w,k) := \frac{1}{k}\sum_{s \in \mathrm{nearest}(w,k)} e(s)
\end{equation}

\textsc{ParaSim} computes the emotion of word $w$ by comparing its similarity with a set of positive and negative paradigm words ($\pos$ and $\neg$, respectively):
\begin{equation}
\label{eq:turney}
e_{\text{\sc ParaSim}}(w) :=\!\!\!\! \sum_{p \in \pos} \!\!\! \sim(w, p) - \!\!\!\! \sum_{n \in \neg} \!\!\!\! \sim(w, n)
\end{equation}
where $\sim(\cdot,\cdot)$ denotes the cosine similarly between two embedding vectors.

Let $e(s)$ map to `$1$', if word $s \in \pos$, and to `$-1$', if $s \in \neg$, then Equation (\ref{eq:turney}) can be rewritten as
\begin{equation}
e_{\text{\sc ParaSim}}(w) =\!\!\!\!\!\!\! \sum_{s \in \pos \cup \neg} \!\!\!\!\!\!\!\!\! \sim(w,s) \times e(s)\text{.}
\end{equation}
For  \textsc{ParaSimNum}, our adaptation of \textsc{Param\-Sim}, we change $e(s)$ to map to a three-dimensional vector corresponding to the VAD entry of a word in our set of seed words $\mathcal{S} := \pos \cup \neg$. We also introduce a normalization factor so that the predictions according to \textsc{ParaSimNum} take the form of a weighted average:
\begin{equation}
e_\text{\sc ParaSimNum}(w) := \frac{\sum_{s \in \mathcal{S}} \sim(w,s) \times e(s)}{\sum_{s \in \mathcal{S}} \sim(w,s)}
\end{equation}

\textsc{RandomWalk} propagates sentiment scores through a graph, with vertices  representing words and edge weights denoting word similarity. 
Let $\mathcal{V}$ represent the set of words in this lexical graph,
and let the vector $p \in \mathbb{R}^{\vert \mathcal{V}\vert}$ represent the induced sentiment score for each word in the graph.
To compute word emotions, $p$ is iteratively updated  by applying a transition matrix $T$:
\begin{equation}
p^{(t+1)} := \beta Tp^{(t)} + (1- \beta)s 
\label{eq:randomwalk_final}
\end{equation}
Here $s \in \mathbb{R}^{\vert \mathcal{V}\vert}$ is the vector representing the seed sentiment scores and the $\beta$-parameter balances between assigning similar scores to neighbors and correct scores to seeds. The vector $p$ is initialized so that the $i$-th element $p_i = 1/|\mathcal{V}|$, whereas $s$ is initialized with $s_i = 1/|\mathcal{S}|$ ($\mathcal{S}$ being the set of seed words), if the corresponding word $w_i$ is a seed word and $0$, otherwise. Details how the transition matrix is initialized can be found in \citet{Zhou04}.

To obtain the final sentiment scores $p_{\mathrm{final}}$, the process is  independently run until convergence for both a positive and a negative seed set, before the resulting values $p^+$ and $p^-$ are normalized by performing a $z$-transformation on:
\begin{equation}
p_{\text{final}} := \frac{p^+}{p^+ + p^-}
\end{equation}

We now provide a simple adaptation for vectorial emotion values, \textsc{RandomWalkNum}: $p$ and $s$ are replaced by $|\mathcal{V}| \! \times 3$ matrices $P$ and $S$, respectively. All entries of $P$ are initialized with $1/|\mathcal{V}|$. For the positive seed set, $S$ is populated with the original VAD values of each word in the seed lexicon and $0$, otherwise. For the negative seed set all values are inverted relative to the center of the numerical VAD rating scales. For instance, the valence score of \textit{relaxed} in Table \ref{tab:examples} is transformed from $7$ to $3$, because $5$ is the center of the respective scale. Finally, $S$ is normalized so that each column adds up to 1. $P_{final}$ can then be calculated analogously to the original algorithm. 
\section{External Datasets} 
\subsection{Diachronic Corpora} 
We rely on two well curated diachronic corpora---the Corpus of Historical American English\footnote{
\myurl{english-corpora.org/coha/}} (COHA; \citealp{Davies12}) and the core corpus of the Deutsches Text Archiv\footnote{ 
\myurl{deutschestextarchiv.de} --- we used the May 2016 snapshot.} ['German Text Archive'] (DTA; \citealp{Geyken13,Geyken15}). They are smaller than some alternative diachronic corpora, especially the Google Books N-gram subcorpora \cite{Lin12}, yet their balanced nature and transparent composition should make results more resilient against \mbox{artifacts} \cite{Pechenick15}. 
Both corpora contain metadata in the form of automatically generated POS annotations and lemmatizations. The latter appears to be more consistent in DTA, possibly due to the inclusion of an orthographic normalization step \citep{Jurish13}. 

COHA is relatively large for a structured corpus \citep[p.\,122]{Davies12} containing over 100k long and short texts from the 1810s to the 2000s. It is conceptually centered around decades and aims at providing equally sized and genre-balanced data for each decade.  The only deviations are an increase in size between the 1810s and 1830s to a then stable level, as well as the inclusion of newspaper texts from the 1860s onwards. COHA is based on post-processed texts from several pre-existing collections, e.g., Project Gutenberg \citep[p.\,125]{Davies12}, digitized with optical character recognition (OCR) software.

DTA is the closest German equivalent to COHA and the result of an ongoing effort to create a digital full-text corpus of printed German documents from the 15\textsuperscript{th} to the 19\textsuperscript{th} century. It is smaller than COHA, containing only about 1.3k long texts, yet of higher quality,  based on extensive manual transcription (mostly double keying, in some cases corrected OCR).
It contains texts from different genres, and individual texts were chosen with an eye toward cultural (not statistical) representativeness. Balance between genres is limited for some timespans, e.g., non-fiction is strongly over-represented in the early 17\textsuperscript{th} century. However, the texts used in our experiments (see below) are well balanced between fictional and non-fictional texts (101 vs.\ 91 texts, respectively).

For both, COHA and DTA, we selected all texts from particular timespans as basis for our experiments. Those timespans served two purposes: (a) when building our gold standard of historical word emotions (Section \ref{sec:gold}) the annotators were requested to rate word emotions according to the respective target language stage; (b) documents associated with the respective timespan were used to train language stage-specific word embeddings (Section \ref{sec:deriving}) in order to model those gold ratings.

The 2000s decade of COHA was an obvious fit for our synchronic experiments in Section \ref{sec:synvad}, as it is the most recent one. For our diachronic experiments in Section \ref{sec:diavad} we aimed at sufficiently sized training material (10M+ tokens) to ensure high quality word embeddings. We also wanted to use data as distant from the present time as possible. We thus picked the 1830s decade of COHA for English and combined thirty years of DTA texts (1810--1839) for German---earlier COHA decades, as well as all individual DTA decades, are of insufficient size. 

\subsection{Emotion Lexicons} 

We now describe the VAD lexicons which were used to provide seed words for both synchronic and diachronic experiments.
Based on its size and popularity, we chose the extended version of \textsc{Anew} (\citealp{Warriner13}; see Section \ref{sec:related}) for English.
Concerning German emotion lexicons, we chose the \textit{Affective Norms for German Sentiment Terms} (\textsc{Angst}; \citealp{Schmidtke14}) which contain 1,003 words and largely follows \textsc{Anew}'s acquisition methodology.

\section{Historical Gold Standard}

\subsection{Dataset Construction}
\label{sec:gold}
In general, native speakers fluent in the respective (sub)language are the only viable option for acquiring a gold standard lexicon of emotional meaning for any language or domain. In the case of historical language older than about a century, this option is off the table due to biological reasons---we simply lack native speakers competent for that specific language period.

As the best conceivable surrogate, we rely on historical language experts for constructing our dataset. The gold standard consists of two parts, an English and a German one, each with 100 words. We recruited three annotators for German and two for English, all doctoral students 
experienced in interpreting 19th century texts.

We selected high-frequency words for the annotation to ensure high quality of the associated word embeddings. The selection was done by, first, extracting adjectives, common nouns and lexical verbs from the 1830s COHA and the 1810--1839 DTA subcorpus and then, second,  randomly sampling 100 words out of the 1000 most frequent ones. We manually excluded two cases of ordinal numerals misclassified as adjectives.

The actual rating process was set up as a questionnaire study following established designs from psychological research \cite{Bradley99,Warriner13}. The participants were requested to put themselves in the position of a person living between 1810 and 1839 for the German data set, or a person living in the 1830s for the English one. They were then presented with stimulus words and used the so-called Self-Assessment Manikin (SAM; \citealp{Bradley94})  to judge the kind of feeling evoked by these lexical items. SAM consists of three individual nine-point scales, one for each VAD dimension. Each of the 27 rating points is illustrated by an cartoon-like anthropomorphic figure serving as a non-verbal description of the scale. Moreover, these figures are supplemented by verbal anchors for the low and high end points of the scales e.g., the rating point ``9'' of the Valence scale  represents ``complete happiness''. They were not provided with or instructed to use any further material or references, e.g., dictionaries.
The final ratings for each word were derived by averaging the individual ratings of the annotators. 

\subsection{Dataset Analysis}
\label{sec:goldana}
\begin{table}[tb]
\center
{\small
\begin{tabular}{ l | c  c  c  c }
\toprule
	&	Valence	&	Arousal	&	Dominance & Mean\\
\midrule
goldEN	&	1.20	&	1.08	&	1.41	&	1.23\\
goldDE	&	1.72	&	1.56	&	2.31	&	1.86\\
Warriner	&	1.68	&	2.30	&	2.16	&	2.05\\
\bottomrule
\end{tabular}
}
\caption{Inter-annotator agreement for our English (goldEN) and German (goldDE) gold standard, as well as the lexicon by \newcite{Warriner13} for comparision; Averaged standard deviation of ratings for each VAD dimension and mean over all dimensions.}
\label{tab:iaa}
\end{table}

We measure inter-annotator agreement (IAA)  by calculating the standard deviation (SD) for each word and dimension and averaging these, first, for each dimension alone, and then over these aggregate values, thus constituting an error-based score (the lower the better). Results are provided in Table \ref{tab:iaa}. In comparison with the lexicon by \newcite{Warriner13}, our gold standard displays higher rating consistency. As average over all three VAD dimensions, our lexicon displays an IAA of 1.23 and 1.86 for English and German, respectively, compared to 2.05 as reported by \newcite{Warriner13}.
This suggests that experts show higher consensus, even when judging word emotions for a historical language period, than crowdworkers for contemporary language. An alternative explanation might be differences in word material, i.e., our random sample of frequent words. 


\begin{table}[t]
\small
    \centering
\begin{tabular}{l|rrr|rrr}
\toprule
{} & \multicolumn{3}{c}{\textbf{historical}} & \multicolumn{3}{c}{\textbf{modern}} \\
{} &    V & A & D & V & A & D \\
\midrule
\textit{daughter} &        3.5 &     4.0 &       4.0 &     6.7 &     5.0 &       5.1 \\
\textit{divine}   &        7.0 &     7.0 &       2.0 &     7.2 &     3.0 &       6.0 \\
\textit{strange}  &        2.0 &     6.5 &       1.0 &     4.7 &     3.5 &       5.3 \\
\bottomrule
\end{tabular}
    \caption{Illustrative example words with large deviation between historical and modern affective meaning; Valence-Arousal-Dominance (VAD) of newly created gold standard compared to \citet{Warriner13}.}
    \label{tab:hist_examples}
\end{table}

Next, we provide a short comparison of historical and modern emotion ratings. This analysis is restricted to the English language, because the overlap of the historical and modern German lexicons is really small (13 words compared to 97 for English). This difference is most likely due to the fact that the English modern lexicon is more than an order of magnitude larger than the German one.

The Pearson correlation between modern and historical lexicons is  0.66, 0.51, and 0.31 for Valence, Arousal, and Dominance, respectively.
Table \ref{tab:hist_examples} displays illustrative examples from our newly created gold standard where historical and modern affective meaning differ strongly.
We conducted a post-facto interview on annotator motivation for those cases. Explanations---which match observations described in common reference textbooks (e.g., \citet{Brinkley03})---range from the influence of feminism leading to an increase in Valence for \textit{``daughter''} up to secularization that might explain a drop in Arousal and rise in Dominance for \textit{``divine''}. The annotation for \textit{``strange''} was motivated by several now obsolete senses indicating foreignness or alienness.\footnote{ See the Oxford English Dictionary: \myurl{oed.com/view/Entry/191244}}

In summary, we 
recruited historical language experts as best conceivable surrogate to compensate for the lack of actual native speakers in order to create a gold standard for historical word emotions. To the best of our knowledge, no comparable dataset is elsewhere available, making this contribution unique and hopefully valuable for future research, despite its obvious size limitation.

\section{Modeling Word Emotions}
\label{sec:experiments}

This section describes how we trained time period-specific word embeddings and used these to evaluate the algorithms presented in Section \ref{subsec:algos} on both a contemporary dataset and our newly created historical gold standard.

\subsection{Word Embedding Training}
\label{sec:deriving}

COHA and DTA were preprocessed by using the lemmatization provided with each corpus, as well as removing punctuation and converting all text to lower case.

We used the \textsc{HyperWords} toolkit \citep{Levy15} to create one distinct word embedding model for each of those subcorpora. Hyper\-para\-meter choices follow \citet{Hamilton16emnlp}. In particular, we trained 300-dimensional word vectors, with a context window of up to four words. Context windows were limited by document boundaries while ignoring sentence boundaries. We modeled words with a minimum token frequency of 10 per subcorpus, different from \citet{Hamilton16emnlp}. For \svdpmi , eigenvectors were discarded, no negative sampling was used and word vectors were combined with their respective context vectors. 

\begin{table*}[tb]
\small
\centering
\begin{tabular}{|c|c|cc|}\hline
Induction Method 				& Seed Selection 	 & \svdpmi & SGNS \\ \hline\hline
\textsc{kNN} 						& full						& \textbf{{0.548}}	& {0.487}\\
\textsc{ParaSimNum} 			& full						& \textbf{{0.557}}	& {0.489}\\
\textsc{RandomWalkNum} 	& full						& \textbf{{0.544}}	& {0.436}\\ \hline
\textsc{kNN} 						& limited 			& {0.181}	& {0.166}\\
\textsc{ParaSimNum}			& limited 			& {0.249}	& {0.191}\\
\textsc{RandomWalkNum}	& limited 			& \textbf{{0.330}}	& {0.181}\\ \hline
\end{tabular}
\caption{Results of the synchronic evaluation in Pearson's $r$ averaged over all three VAD dimensions. The best system for each seed lexicon and those with statistically non-significant differences (p $\geq$ 0.05) are in \textbf{bold}.\label{tab:synchron}}
\vspace*{\baselineskip}
\begin{tabular}{|c|c|c|cc|}\hline
Language	&	Induction Method 		& Seed Selection 	& \svdpmi & SGNS \\ \hline\hline
\parbox[t]{2mm}{\multirow{6}{*}{\rotatebox[origin=c]{90}{English}}}
			&	\textsc{kNN} 						& full			& {0.307}	& \textbf{{0.365}}	\\ 
			&	\textsc{ParaSimNum} 			& full				& {0.348}	& {0.361}	\\ 
			&	\textsc{RandomWalkNum} 	& full				& {0.351}	& {0.361}	\\ \cline{2-5}
			&	\textsc{kNN} 						& limited 		& {0.273}	& {0.153}\\
			&	\textsc{ParaSimNum}			& limited 		& {0.295}	& {0.232}\\
			&	\textsc{RandomWalkNum}	& limited 		&\textbf{ {0.305}}	& {\ \ \ 0.039$^\triangle$}\\ 
			\hline\hline
\parbox[t]{2mm}{\multirow{3}{*}{\rotatebox[origin=c]{90}{German}}}
			&	\textsc{kNN} 						& full				& {0.366}	& {0.263}\\
			&	\textsc{ParaSimNum} 			& full				& \textbf{{0.384}}	& {0.214}\\
			&	\textsc{RandomWalkNum} 	& full				& {0.302}	& {0.273}\\  \hline
\end{tabular}
\caption{Results of the diachronic evaluation in Pearson's $r$ averaged over all three VAD dimensions. The best system for each language and seed selection strategy (\textit{full} vs. \textit{limited}) is in \textbf{bold}. Only the system marked with `$^\triangle$' is significantly different from the best system (p $<$ 0.05).}\label{tab:diachron}
\end{table*}

\subsection{Synchronic Evaluation}
\label{sec:synvad}

Our first evaluation of lexicon induction algorithms compares the ability of the three different algorithms described in Section \ref{sec:algorithms} to predict ratings of a modern, contemporary VAD lexicon, i.e., the one by \citet{Warriner13}, using two different types of seed sets (see below). For this experiment, we used word embeddings trained on the 2000s COHA subcorpus. We call this evaluation setup \textit{synchronic} in the linguistic sense, since seed lexicon, target lexicon and word embeddings belong to the same language period. A unique feature of our work here is that we also take into account possible interaction effects between lexicon induction algorithms and word embedding algorithms, i.e., SGNS and \svdpmi . 

We use two different seed lexicons, both are based on the word ratings by \newcite{Warriner13}. The \textit{full} seed lexicon corresponds to all the entries of words which are also present in \textsc{Anew} (about 1,000 words; see Section \ref{sec:related}). In contrast, the \textit{limited} seed lexicon is restricted to 19 words\footnote{
One of the 20 words given by \newcite{Hamilton16emnlp}, \textit{``hated''}, is not present in the Warriner lexicon and was therefore eliminated.} which were identified as temporally stable by \newcite{Hamilton16emnlp}.

The first setup is thus analogous to the polarity experiments performed by \newcite{Cook10}, whereas the second one corresponds to the settings from \newcite{Hamilton16emnlp}. We use Pearson's $r$ between actual and predicted values for each emotion dimension (Valence, Arousal and Dominance) for quantifying performance\footnote{
  Some other studies use the rank correlation coefficient Kendall's $\tau$. We found that for our experiments the results are overall consistent between both metrics. In the following we only report Pearson's $r$ as it is specifically designed for numerical values. In contrast, Kendall's $\tau$ only captures ordinal information and is therefore less suited for VAD.
} and a Fisher transformation followed by a Z-test for significance testing \citep[pp. 130--131]{Cohen95}.

Table \ref{tab:synchron} provides the average values of these VAD correlations for each seed lexicon, embedding method and induction algorithm. SGNS embeddings are worse than \svdpmi \ embeddings for both full and limited seed lexicons. \svdpmi \ embeddings seem to be better suited for induction based on the full seed set, leading to the highest observed correlation with \textsc{ParaSimNum}. However, results with other induction algorithms are not significantly different. For the limited seed set, consistent with claims by \citet{Hamilton16emnlp}, \textsc{RandomWalkNum} is significantly better than all alternative approaches. However, all results with the limited seed set are far (and significantly) worse than those with the full seed lexicon.

Performance is known to differ between VAD dimensions, i.e., Valence is usually the easiest one to predict. For the full seed lexicon and the best induction method, \textsc{ParaSimNum} with \svdpmi\ embeddings, we found Pearson's $r$ correlation 
to range between 0.679 for Valence, 0.445 for Arousal and 0.547 for Dominance.

\subsection{Diachronic Evaluation}
\label{sec:diavad}

The second evaluation set-up utilizes our historical gold standard described in Section \ref{sec:gold}. 
We call this set-up \textit{diachronic}, since the emotion lexicons generated in our experiments aim to match word use of \textit{historical} language stages, whereas the seed values used for this process stem from \textit{contemporary} language. 
This approach allows us to test the recent claim that artificially \textit{limiting} seed lexicons to words assumed to be semantically stable over long time spans is  beneficial for generating historical emotion lexicons \cite{Hamilton16emnlp}. We used Pearson's $r$ correlation and the Z-test, as in Section~\ref{sec:synvad}.

Again, we investigate interactions between lexicon induction algorithms and embedding types. For English, we evaluate with both \textit{full} and \textit{limited} seed lexicons, whereas for German, we evaluate only using the \textit{full} seed lexicon (\textsc{Angst}, see Section \ref{sec:related}) since most entries of the English \textit{limited} lexicon have no corresponding entry in \textsc{Angst}. Embeddings are based on the 1830s COHA subcorpus for English and on  the 1810--1839 DTA subcorpus for German, thus matching the time frames featured by our gold standard. 

The results of this experiment are given in Table \ref{tab:diachron}. For English, using the full seed lexicons, we achieve performance figures around $r=.35$. In contrast, using the \textit{limited} seed lexicon we find that the performance is markedly weaker in each of our six  conditions compared to using the full seed lexicon. 
This observation directly opposes the claims from \citet{Hamilton16emnlp} who argued that their hand selected set of emotionally stable seed words would boost performance relative to using the full, contemporary dataset as seeds. 

Our finding is statistically significant in only one of all cases (the combination of  SGNS and \textsc{RandomWalkNum}). However, the fact that we get the \textit{identical} outcomes for all the other five combinations of embedding and induction algorithm strongly indicates that using the full seed set is virtually superior, even though the differences are not statistically significant when looking at the individual conditions in isolation, due to the size\footnote{
    Typical emotion lexicons are one or even two orders of magnitude larger, as discussed in Section \ref{sec:representing}. Given the current correlation values, we would need to increase the size of our gold standard by a factor of about 40---a challenging task, given its expert reliant nature---to ensure $p < .05$.} 
of our gold standard. Note that this  outcome is also consistent with our results from the synchronic evaluation where we did find significant differences.


German results with the full seed lexicon are similar to those for English. Here, however, the SGNS embeddings are outperformed by \svdpmi , whereas for English both are competitive. A possible explanation for this result might be 
differences in pre-processing between the two data sets which were necessary due to the more complex morphology of the German language. 

\section{Conclusion}
In this contribution, we addressed the task of constructing emotion lexicons for historical language stages. We presented adaptations of two existing polarity lexicon induction algorithms to the multidimensional VAD model of emotion, which provides deeper insights than common bi-polar approaches.  Furthermore, we  constructed the first gold standard for affective lexical semantics in historical language. In our experiments, we investigated the interaction between word embedding algorithm, word emotion induction algorithm and seed word selection strategy. Most importantly, our results suggest that limiting seed words to supposedly temporally stable ones does not improve performance as suggested in previous work but rather turns out to be  harmful.
Regarding the compared algorithms for emotion lexicon induction and embedding generation, we recommend using \svdpmi\ together with \textsc{ParaSimNum} (our adaption of the \citet{Turney03} algorithm), as this set-up yields strong and stable performance, and requires few hyperparameter choices. 
We will continue to work on further solutions to get around data sparsity issues when working with historical language, hopefully allowing for more advanced machine learning approaches in the near future. 

\section*{Acknowledgments}
We thank our emotion gold standard annotators for volunteering. 
This research was partially funded by the Deutsche Forschungsgemeinschaft (DFG) within the Graduate School \textit{The Romantic Model} (GRK 2041/1).
\bibliographystyle{acl_natbib}
\bibliography{paper}

\end{document}